\documentclass{article}

     \usepackage[preprint]{neurips_2022}

\usepackage[utf8]{inputenc} 
\usepackage[T1]{fontenc}    
\usepackage{hyperref}       
\usepackage{url}            
\usepackage{booktabs}       
\usepackage{amsfonts}       
\usepackage{nicefrac}       
\usepackage{microtype}      
\usepackage{xcolor}         

\usepackage{amsmath}
\usepackage{amssymb}
\usepackage{bbm}
\usepackage{algorithm}
\usepackage{algpseudocode}
\usepackage{graphicx}

\title{Bandits for Online Calibration: An Application to Content Moderation on Social Media Platforms}

\author{\textbf{Vashist Avadhanula$^{\diamond,1}$, Omar Abdul Baki$^{\diamond}$, Hamsa Bastani$^{\diamond,\dagger,2}$, Osbert Bastani$^{\diamond,\dagger,3}$, Caner} \\ \textbf{Gocmen$^{\diamond}$,   Daniel Haimovich$^{\diamond}$, Darren Hwang$^{\diamond}$, Dima Karamshuk$^{\diamond}$ ,  Thomas Leeper$^{\diamond}$,} \\  \textbf{ Jiayuan Ma$^{\diamond}$,  Gregory  Macnamara$^{\diamond}$, Jake Mullett$^{\diamond}$, Christopher Palow$^{\diamond}$,  Sung Park$^{\diamond}$,} \\ \textbf{ Varun S Rajagopal$^{\diamond}$, Kevin Schaeffer$^{\diamond}$, Parikshit Shah$^{\diamond}$, Deeksha Sinha$^{\diamond}$,  Nicolas } \\ \textbf{ Stier-Moses$^{\diamond}$, Peng Xu $^{\diamond}$} \\ \\
  $^{\diamond}$Meta, $^\dagger$University of Pennsylvania
}

\begin{document}

\maketitle

\begin{abstract}
We describe the current content moderation strategy employed by Meta to remove policy-violating content from its platforms. Meta relies on both handcrafted and learned risk models to flag potentially violating content for human review. Our approach aggregates these risk models into a single ranking score, calibrating them to prioritize more reliable risk models. A key challenge is that violation trends change over time, affecting which risk models are most reliable. Our system additionally handles production challenges such as changing risk models and novel risk models. We use a contextual bandit to update the calibration in response to such trends. Our approach increases Meta's top-line metric for measuring the effectiveness of its content moderation strategy by 13\%.
\end{abstract}
\section{Introduction}

Meta has nearly 3.71 billion (monthly) active users worldwide, with billions of pieces of content shared by users every day \citep{metaearnings}. While most content is benign, a small share---e.g., hate speech, promoting terrorism, or graphic pornography---violates the platform's community standards. Thus, a key goal is to promptly remove such content.

\begin{figure}
\centering
\includegraphics[width=0.95\textwidth]{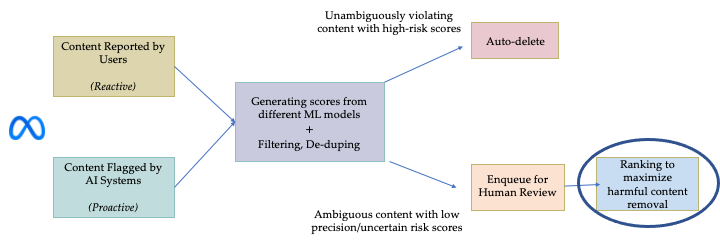}
\caption{Decision Flow Diagram for Content Moderation. This paper describes how we optimize reviewer capacity by prioritizing within the subset of potentially policy-violating content flagged by many risk models (circled in blue).}
\label{fig:flow}
\end{figure}

Determining whether a piece of content is policy-violating is a difficult decision that often requires manual review, making it challenging to scale to Meta’s prolific content streams. As of December 2021, over 2 million pieces of content are reviewed per day by around 15,000 human reviewers across the globe~\citep{fbspam}. Clearly, manually reviewing all content is infeasible; thus, Meta relies heavily on risk models (based on machine learning or handcrafted rules) to flag potentially violating content pieces. A subset of this content is deemed unambiguously violating and is automatically removed; the remainder undergoes manual review (see Fig.~\ref{fig:flow}).

\begin{figure}
\centering
\includegraphics[width=.8\columnwidth]{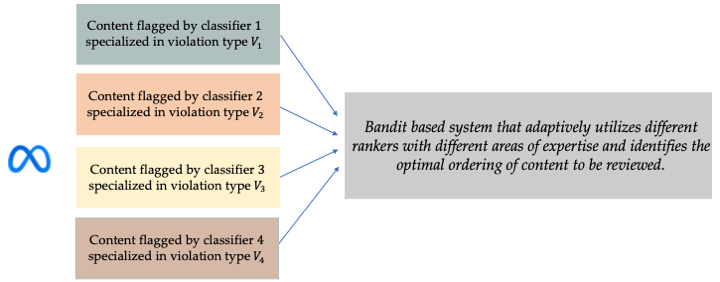}
\caption{Centralized Bandit Approach for Content Moderation. The bandit algorithm dynamically aggregates the outputs of many different violation-specific risk models to learn a prioritization ranking that maximizes integrity value.}
\label{fig:bandit}
\end{figure}

To maximize the amount of violating content removed given a fixed supply of human reviewers, we dynamically prioritize content likely to have wide reach and high severity (quantified by a metric called \emph{integrity value}; see Section~\ref{sec:problem}). Initial efforts designed separate risk scores for different content violation types (e.g., one for violations of the nudity policy, one for hate speech, etc.), allocating a fixed, pre-determined reviewer capacity to each type~\citep{wpie}. However, this strategy was unable to adapt to the nonstationary and quickly varying violation trends---i.e., violating content constantly changes in appearance, focus, and wording. Exacerbating this issue, users trying to post violating content are adversarial, attempting to evade detection. Thus, the performances of the different risk models are constantly changing. Models require significant expertise and effort to retrain and can only be updated periodically, and new risk models are frequently added for emerging trends---e.g., to capture the use of emojis in discussions of English football to detect a recent wave of racist comments directed at black football players \citep{BBC}.

Thus, Meta moved to a single holistic ranker ``Whole Post Integrity Embeddings'' (WPIE), a pretrained universal representation of content across modalities and violation types~\citep{ wpie, halevy2022preserving}. It resulted in significant performance improvements in production~\citep{wpie, wpieperf}, and is the baseline our work improves upon. An important drawback is that, although more accurate overall, it lacked precision for specific types of violations compared to the risk model tailored to that type. Furthermore, while it was periodically retrained to handle evolving trends, it did not incorporate sufficient exploration of new violation types to ensure sufficient responsiveness to new trends.

To address these challenges, we have designed and deployed a bandit approach that combines the full set of available risk models into a single risk score; importantly, it \emph{calibrates} the scores to prioritize ones with higher quality, doing so \emph{dynamically} to respond to new trends  (see Fig.~\ref{fig:bandit}). A key challenge is \textit{bandit feedback}: we only observe the true severity of a piece of content when it is reviewed by a human. Thus, we must actively explore different types of content to obtain ground truth data on the accuracy of different risk models. Finally, the scale of the problem imposes constraints on the techniques we can leverage---we must ensure that our system can assess a large number of content every minute for potential violations under computational cost and latency constraints.

In detail, we use a nonstationary, batched contextual bandit that treats the various risk models as features for predicting severity. We address two key production-related challenges that differ from existing approaches. First, for a given piece of content, our prediction of its reach (and therefore potential prevalence) evolves over time as it is viewed and shared by users. However, traditional bandit algorithms assume that features are static. We incorporate the bandit into a queuing framework, where the queue priority is determined by both the predicted severity as well as the velocity of its predicted reach. Second, we must incorporate new risk models (i.e., features) seamlessly despite the fact that we cannot retrain our severity predictions online due to production constraints. Together, our approach effectively learns an estimate of severity that dynamically adapts to changes in the environment despite production constraints. Our approach has been deployed at scale at Meta, increasing the top-line metric of Integrity Value (which quantifies the impact of removing violating content accounting for severity and potential reach) by 13\% compared to the previous WPIE approach.

\paragraph{Related Work.}

AI has become a promising strategy for ensuring content integrity~\citep{halevy2022preserving}---e.g., natural language processing is widely used to identify abusive language, hate speech, and cyber-bullying~\citep{nobata2016abusive, zhong2016content, hosseinmardi2015detection, djuric2015hate, burnap2016us, chen2012detecting, gamback2017using, ma2016detecting, van2018automatic, noorshams2020ties}. Our work addresses this problem using contextual bandits, which has previously been applied in settings such as news article recommendations~\citep{li2010contextual} and personalized healthcare~\citep{bastani2020online,tewari2017ads}. Building on the Upper Confidence Bound (UCB) algorithm~\citep{auer2010ucb, chu2011contextual}, our bandit incorporates techniques to handle nonstationarity~\citep{besbes2014stochastic} and batched predictions~\citep{perchet2016batched,gao2019batched}. We also address several novel practical challenges. First, we propose a simple approach to compute uncertainty online, dramatically improving scalability. Second, our approach can dynamically incorporate new risk models that increase the dimension of our contexts.


\section{Problem Formulation}
\label{sec:problem}

\paragraph{Content arrival process.}

We assume content arrives sequentially. At each step $t$, content $c_t$ arrives, and we observe its features $x_t=\phi(c_t) \in \mathbb{R}^d$; each feature is the output of a model that predicts the real-valued risk of $c_t$ for a given violation type (e.g., hate speech). These features are built by various teams at Meta, and range from machine learning algorithms trained on reviewers’ historical labels, to simple classifiers based on regular expressions to flag violating phrases. Each piece of content receives $d$ predicted risk scores (from each of these models), which are then concatenated to form $x_t$.

\paragraph{Manual review decisions.}

Next, our system must decide whether to have a reviewer label $c_t$. We formalize this process as a one-armed bandit~\citep{woodroofe1979one}, where the arm is the action
\[a_t=\mathbbm{1}(\text{mark }c_t\text{ for manual review}),\]
and $\mathbbm{1}$ is the indicator function. That is, pulling the arm corresponds to having a reviewer manually examine $c_t$, and not pulling it corresponds to leaving the content up without review.

\paragraph{Objective.}

There are several desiderata when assessing the risk of violating content. For instance, we wish to prioritize content that has a high likelihood of violation, but we also wish to prioritize content with more severe violations (e.g., terrorism or child nudity). Additionally, we wish to prioritize content that is likely to receive a large number of views. We use the \emph{Integrity Value (IV)}, which quantifies the value of taking down a piece of violating content, as our key metric of interest. At a high level, IV of a piece of content has the form
\[ IV= (\text{predicted future views} + \text{constant}) \times (\text{severity}) .\]
The additive constant is tuned to sufficiently prioritize nonviral but violating content with high severity. Future views are predicted dynamically using past viewership trajectories of similar content by similar users. We fit a Hawkes process, which is effective at capturing ``self-exciting'' phenomena such as viral content on social media~\citep{haimovich2022popularity}. Severity is a real-valued function that maps every possible policy violation type to a non-negative real number. The system level IV is defined as the sum of the IV of all the content sent for human review.

Once a piece of content is flagged for human review (i.e. $a_t=1$), a human reviewer determines its severity $y_t$. We define $y_t=0$ for non-violating content. If $y_t>0$, then we remove the underlying content and reward the underlying risk model (see Section~\ref{sec:algorithm}). 

\paragraph{Practical challenges.}

We briefly describe several practical challenges our algorithm must deal with; see Section~\ref{sec:implementation} for details. First, we must handle \emph{content lifetime}---rather than make an immediate decision, we can defer the decision to a future time step (e.g., when more reviewers are available); however, there is a cost of leaving violating content up for a longer period of time. Second, since the algorithm must run in real-time on a huge volume of content, it cannot involve overly complex computations. Third, we must handle \emph{nonstationarity}---violating content changes form over time, meaning older data may not be representative of the current content.

\section{Bandit Algorithm}
\label{sec:algorithm}


\paragraph{Risk prediction model.}

We describe our algorithm for deciding whether to flag each $c_t$ for review. Na\"{i}vely, we could flag $c_t$ if any of its risk scores $x_{t,i}$ are positive---i.e., construct an overall risk score
$\hat{y}_t=\max_ix_{t,i}$ and take $a_t=\mathbbm{1}(\hat{y}_t>0)$ (our implementation uses the predicted risk scores to prioritize content rather than making isolated decisions; see Section~\ref{sec:implementation}). However, different risk scores may not be directly comparable. To address this problem, our algorithm rescales the different risk scores according to some parametric function $f_\beta$, which we refer to as \emph{calibrating} the set of risk models:
\begin{align*}
\hat{y}_t=\max_if_{\beta_i}(x_{t,i}),
\end{align*}
where $\beta_i\in\mathbb{R}^k$ are rescaling parameters for risk model $i$. We choose $f_{\beta}$ to be piecewise linear:
\begin{equation}\label{eq:piece_wise_linear}
f_{\beta_i}(z)=\sum_{j=1}^k\mathbbm{1}(z\in B_j)\beta_{i,j}z
\end{equation}
for each $i$, where $B_j\subseteq\mathbb{R}$ are bins over the space of risk scores. For simplicity, we have assumed that the bins are identical across different risk scores $i$, but in our implementation, they actually depend on $i$; roughly speaking, they are chosen based on quantiles of the observed scores $\{x_{t,i}\}_t$. Our goal is to learn the unknown calibration parameters $\{\beta_i\}$ to make effective labeling decisions.

\paragraph{Parameter estimation.}

Next, we describe how to estimate $\beta_i$ given a fixed dataset $\{(x_t,y_t)\}_t$. In this setting, we can estimate $\beta_{i,j}$ for each risk score $i$ and bin $B_j$ independently, since it is a linear model:
\begin{equation}\label{eq:linear_model}
y_t=\beta_{i,j} x_{t,j}+\epsilon_{t,i,j},
\end{equation}
for some $\sigma$-subgaussian noise term $\epsilon_{t,i,j}$. Thus, we can estimate $\beta_{i,j}$ using linear regression:
\begin{align}
\label{eq:beta_update}
\hat\beta_{i,j}=\Big(\sum_tx_{t,j}^2\Big)^{-1}\Big(\sum_tx_{t,j}y_t\Big).
\end{align}
In addition, our implementation uses a heuristic where for the parameters $\beta_i$ for risk score $i$, we only train on a top $\alpha$ quantile of examples $\{x_{t,i}\}_t$ in terms of magnitude, where $\alpha$ is a hyperparameter. Importantly, this strategy only depends on the relative magnitude of risk scores given by scoring function $i$, so it is not affected by the fact that different risk scores $x_{t,i}$ and $x_{t,i'}$ are incomparable.

\paragraph{Upper confidence bounds.}

A key challenge is \textit{bandit feedback}: we only observe the true severity $y_t$ for content $c_t$ when $a_t=1$. Thus, we must actively explore different types of content to obtain labels to estimate $\beta_i$. Intuitively, we mark content for review either when its label can provide information towards better estimating some $\beta_i$ (exploration), or when it is likley to be violating (exploitation). In particular, we use UCB~\citep{auer2010ucb}, which chooses arms based on an \textit{optimistic} estimate of its rewards. It maintains both a point estimate $\hat{\beta}_{i,j}$ of $\beta_{i,j}$, and an upper confidence bound for this estimate:
\begin{equation}
\label{eq:ucb_def}
\mathbb{P}[\hat{\beta}_{i,j}+u_{i,j}\ge\beta_{i,j}]\ge1-\delta,
\end{equation}
where $\beta_{i,j}$ are the ``true'' parameters, $\delta\in(0,1)$ is a confidence level, and the probability is taken with respect to the randomness in the examples used to train $\hat\beta_{i,j}$. For our model, we can take
\begin{equation}
\label{eq:ucb_bias}
u_{ij}=\sigma_{i,j} \cdot \sqrt{\frac{\log(1/\delta)}{\sum_tx_{t,i}^2\mathbbm{1}(x_{t,i}\in B_j)}} .
\end{equation}
The denominator is an estimate of the covariance restricted to samples in bin $B_j$, and the numerator scales the bound to provide a high-probability guarantee. Finally, $\sigma_{i,j}$ captures the noise variance; in practice, it is unknown, so we estimate it as the empirical standard error of our linear regression estimate.
Also, as we rely on a fixed window of training data, we use a fixed choice of $\delta$ (whereas standard bandit algorithms reduce $\delta$ over time).  Finally, for content with features $x_i$, we \textit{optimistically} estimate its severity to be
\begin{equation}
\label{eq:final_optimistic_estimate}
\hat y_t=\max_if_{\hat\beta_i+u_i}(x_{t,i}).
\end{equation}
Our full algorithm is summarized in Algorithm \ref{alg:bandit}.

\begin{algorithm}[t]
\caption{Basic Bandit Algorithm}\label{alg:bandit}
\begin{algorithmic}
\State Initialize parameters $\hat\beta_i$, uncertainties $u_i$ 
\For{$t\in\mathbb{N}$}
\State Observe content risk scores $x_t=\phi(c_t)$
\State Choose action $a_t=\mathbbm{1}(\hat y_t>0)$, where $\hat y_t = \max_i f_{\hat\beta_i+u_i}(x_{t,i})$
\If{$a_t=1$}
\State Mark $c_t$ for review, and observe true severity $y_t$
\State Update parameters $\hat\beta_i$ using \eqref{eq:beta_update}, uncertainties $u_i$ using \eqref{eq:ucb_bias}
\EndIf
\EndFor
\end{algorithmic}
\end{algorithm}

\section{Implementation Challenges}
\label{sec:implementation}

\paragraph{Content lifetime.}

A key challenge is that content actually persists on the platform for an extended period of time. Thus, we can revisit negative decisions $a_t=0$ at future steps---e.g., reviewers may become less busy, making it worthwhile to review the content; alternatively, the IV of the content may increase since it scales with viewership, which is time-varying. Rather than make a binary decision, our algorithm instead maintains a pool $C_t$ of all currently available content on time step $t$. In this formulation, time steps correspond to events where a reviewer becomes available (rather than where a new content arrives on the platform)---e.g., multiple pieces of content may be added and/or removed from $C_{t-1}$ to obtain $C_t$. Also, note that the action space is now the content $c_t^*$ to review on step $t$ rather than whether to review content $c_t$. Na\"{i}vely, we could review content with highest predicted severity/IV:
\begin{align*}
c_t^*=\max_{c\in C_t}\{\max_if_{\hat\beta_i+u_i}(\phi(c))\},
\end{align*}
where the term inside the brackets is the predicted severity of $c$. However, a key insight from the literature on job scheduling is that the optimal policy allocates jobs with the highest \emph{rate of change} first, known as the \emph{$c\mu$ rule}~\citep{cmu}. Adapting this rule to our setting, we prioritize content based on its estimated \emph{rate of change} in IV.

\paragraph{Real-time parameter updates.}

Meta evaluates a very large volume of content for potential violations every minute. To ensure that we can dynamically adapt to new violation trends, we update our parameter estimates once every 5 minutes. For scalability, we use online updates to our parameter estimates rather than re-computing them from scratch. Specifically, let $XY^t_{ij} = \sum_{t} y_t x_{ti} \mathbbm{1}(x_{ti}  \in B_j)$ and $XX^t_{ij} = \sum_t x^2_{ti} \mathbbm{1}(x_{ti}  \in B_j)$ be the running numerators and denominators computed until time $t$ in \eqref{eq:linear_model}. We can compute the terms $XY^t_{ij}$ and $XX^t_{ij}$ online as follows:
\begin{equation}
\begin{aligned}
XY^{t+1}_{ij} &= XY^t_{ij} + x_{t+1,i}y_{t+1},
\qquad
XX^{t+1}_{ij} = XX^t_{ij} + x^2_{t+1,i}.
\end{aligned}
\end{equation}
Therefore, $\beta_{ij}^{t+1}$ can be computed online as the ratio of $XY^{t+1}_{ij}$ and $XX^{t+1}_{ij}$. Next, let $N_{ij}^t = \sum_{\tau<t} \mathbbm{1}(x_{\tau,i} \in B_j)$ be the number of labeled content pieces until time $t$ by ranker $i$ that belonged to bin $j$. Then, we can compute the standard errors online as follows:
\begin{align*}
(\sigma_{ij}^{t+1})^2 = \frac{N_{ij}^t}{N_{ij}^{t+1}}\Big((\sigma_{ij}^t)^2 + y^2_{t+1} + (\beta_{ij}^t)^2 XX^t_{ij} - (\beta_{ij}^{t+1})^2 XX^{t+1}_{ij}\Big)
\end{align*}
This strategy reduces the number of queries to our database from tens of thousands to a single look-up.

\paragraph{Nonstationarity.}

An important challenge is that violation trends are highly non-stationary, since users may learn to evade detection, and since Meta regularly updates its community standards. Thus, we exponentially downweight older, less representative content: given a discount factor $\gamma$, we weight training examples by $\gamma^\tau$, where $\tau$ is the number of hours since the content arrived. We compute the reweighted variants of $XX^t_{ij}$ and $XY^t_{ij}$ as follows:
\begin{equation*}
\begin{aligned}
XY^t_{ij} &= \sum_{\tau} \gamma^{t-\tau} x_{\tau,i} y_{\tau},\qquad
XX^t_{ij} &= \sum_{\tau} \gamma^{t-\tau} x^2{\tau,i}.
\end{aligned}
\end{equation*}
Similarly, we compute $\sigma_{ij}^t$ and $u_ij$ as follows:
\begin{equation*}
\begin{aligned}
\sigma_{ij}^t &= \sqrt{\frac{\sum_{\tau} \gamma^{2{t-\tau}}(y_{\tau}-\hat{\beta}_{ij}^tx_{\tau, i})^2}{\sum_\tau \gamma^{t-\tau}\mathbbm{1}(x_{\tau,i} \in B_j)}},
\qquad
u^t_{ij} = \sigma_{ij}^t \cdot \sqrt{\frac{\log(1/\delta)}{XX^t_{ij}}} .
\end{aligned}
\end{equation*}
Finally, we altogether remove content beyond a window $\tau\ge\tau_{\text{max}}$. 

\paragraph{Adding new features.}

Another way to address nonstationarity is to add new features (i.e., risk models). One example is the recent use of emojis in a wave of racist comments directed at black football players immediately after the UEFA European Football Championship final~\citep{BBC}. Here, the key indicator of violating content was the use of a particular set of derogatory emojis (which were benign outside this context). Retraining existing risk models to identify such new trends is often time-consuming, slowing down our response to fast-moving violation trends. Instead, we quickly deployed a specialized risk model that flags any content including both these emojis and football discussion, combining it with sentiment analysis to estimate the violation likelihood. We have developed infrastructure to quickly launch such handcrafted rankers. In particular, our bandit algorithm quickly learns the effectiveness of this new risk model via exploration. If a violation trend is short-lived, it will also quickly learn that the new risk model's utility has decreased, and will downweight it accordingly. Subsequently, we have observed substantial improvements in the turnaround time for flagging and removing new variants of violating content.

\paragraph{Interpretability.}

A key advantage of our approach is interpretability: it is easy to understand why certain piece of content is flagged for review, since we can point to the risk model responsible for high severity. This ability helps debug potential issues when there are unexpected increases in a certain type of violating content.

\section{Deployment}
\label{ssec:deployment-details}

\begin{figure}[htb]
\begin{center}
\includegraphics[width=0.50\textwidth]{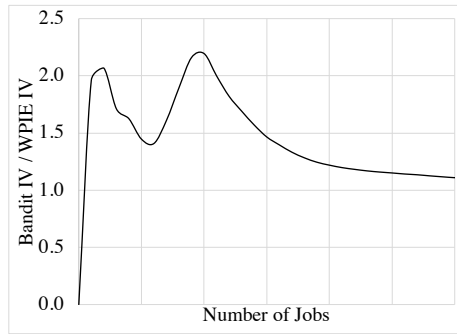}
\caption{Relative performance of the Centralized Bandit approach}
\label{fig:sim_iv_gain}
\end{center}
\end{figure}

Our approach has been deployed at scale at Meta. One challenge was how to choose hyperparameters, which can significantly affect performance~\citep{bietti2018practical}. To do so, we developed a simulator based on a rolling 30 days of historical data.
We chose UCB (which outperformed Thompson Sampling) based on the simulator, as well as $\delta$ (the confidence parameter) and $\gamma$ (the discount factor). Further, using simulations we established the significant IV gains that our approach would have as compared to the existing approach in production for a large range of number of jobs processed (exact scale of number of jobs has been anonymized) as shown in Figure \ref{fig:sim_iv_gain}.

Internal A/B tests were run between 9th August, 2021 and 20th August, 2021, spanning a large number of jobs and reviewer hours. Further, the experiment was run across multiple language based markets including Indonesian, Turkish, Portuguese, German, Thai,  Korean, Mexican, Hindi, Japanese, Romanian, Dari, Filipino, Urdu, Burmese, Pashto, Hebrew and region based markets including Australia and North America. There were three markets we removed from the test midway due to world events (Dari, Pashto and Urdu). Results in these markets were hence excluded. The A/B test consisted of 1,184,526 jobs being reviewed based on the control algorithm and 963,908 jobs being reviewed based on the bandit algorithm.

In these tests, the bandit approach was compared to the existing approach used in production before ours, called \emph{Whole Post Integrity Embeddings (WPIE)}~\citep{wpie}. WPIE uses a single ranker across different violation types based on a pretrained universal representation of content for integrity problems; WPIE itself significantly improved IV compared to an initial strategy that used fixed allocations for different risk models~\citep{wpieperf}.

These tests have demonstrated a statistically significant 13\% (0.9\%, 25.2\%) lift in IV (with a fixed capacity of human reviewers) compared to the WPIE (The values in the brackets indicate 95\% confidence intervals).  To maintain the same IV as our bandit, WPIE would need approximately 780,000 additional people-hours per year of human reviewer capacity.  Finally, an important advantage of our approach is the ability to seamlessly handle new features that capture novel violation trends.


\section{Conclusion}

We have described our system deployed at Meta for identifying and removing violating content from the platform. Our system employs a bandit algorithm to dynamically adjust calibration parameters across a set of risk models. In an internal A/B test, our system outperformed the existing approach, WPIE, by over 13\% in terms of IV with a fixed capacity of human reviewers. As of December 2021, our system flags over 2 million pieces of content for review per day by over 15,000 human reviewers across the globe~\citep{fbpolicy}. Our work demonstrates that bandit algorithms are a promising strategy for addressing issues in deploying machine learning systems in highly nonstationary environments.

\bibliographystyle{unsrt}
\bibliography{refs}

\appendix

\end{document}